\documentclass[letterpaper,10pt,conference]{ieeeconf}  

\IEEEoverridecommandlockouts                              
\usepackage{graphicx} 
\usepackage{amsmath} 
\usepackage{amssymb}  
\usepackage{algorithm}
\usepackage{algorithmic}
\usepackage{xcolor}

\makeatletter
\let\NAT@parse\undefined
\makeatother
\usepackage{hyperref}
\usepackage{float}
\graphicspath{{figures/}}

\title{\LARGE \bf Task Planning with a Weighted Functional Object-Oriented Network}

\author{David Paulius, Kelvin Sheng Pei Dong, and Yu Sun\\
\thanks{Yu Sun leads the Robot Perception and Action Lab (RPAL), which is a part of the Department of Computer Science \& Engineering at the University of South Florida, Tampa, FL, USA. David and Kelvin were formerly student researchers at RPAL and are now posted at the Technical University of Munich and University of Illinois - Urbana Champaign respectively.
 \newline(Contact email: \texttt{david.paulius@tum.de,yusun@usf.edu)}}%
}

\begin{document}

\maketitle

\thispagestyle{empty}
\pagestyle{empty}

\begin{abstract}
In reality, there is still much to be done for robots to be able to perform manipulation actions with full autonomy. Complicated manipulation tasks, such as cooking, may still require a person to perform some actions that are very risky for a robot to perform. On the other hand, some other actions may be very risky for a human with physical disabilities to perform. Therefore, it is necessary to balance the workload of a robot and a human based on their limitations while minimizing the effort needed from a human in a collaborative robot (cobot) set-up. This paper proposes a new version of our \emph{functional object-oriented network} (FOON) that integrates weights in its functional units to reflect a robot's chance of successfully executing an action of that functional unit. The paper also presents a task planning algorithm for the weighted FOON to allocate manipulation action load to the robot and human to achieve optimal performance while minimizing human effort. Through a number of experiments, this paper shows several successful cases in which using the proposed weighted FOON and the task planning algorithm allow a robot and a human to successfully complete complicated tasks together with higher success rates than a robot doing them alone.
\end{abstract}

\section{Introduction}

A major prospect in robotics is to develop robots that are programmed with the ability to perform daily tasks autonomously, which would particularly be useful for persons who cannot perform them on their own.
Such robots would rely on several components, such as perception modules, programmed motion primitives and skills, and knowledge representation, to understand the effects of its actions on the state of its environment~\cite{paulius2019survey}.
Inspired by prior work~\cite{Ren2013,SunRAS2013,Lin2015a}, we introduced the functional object-oriented network (FOON) as a knowledge representation for service robots~\cite{paulius2016functional,paulius2018functional}.
FOON, which is motivated by the theory of affordance~\cite{Gibson_1977}, describes the relationship between objects and manipulation actions as nodes in the network.
In~\cite{paulius2016functional}, we showed how knowledge can be merged into a single network from which knowledge can be retrieved as task sequences, while in~\cite{paulius2018functional}, we demonstrated how existing knowledge can be used to learn ``new'' concepts based on object similarity.

However, up to this point, we have yet to demonstrate how a real robot can use FOON for task execution, as to perfectly design a robot that can understand its actions and operate safely in human environments is an exceptionally daunting task.
Firstly, the robot's working space (i.e. human environments) is usually very dynamic, as it is likely to feature objects of different shapes, sizes, positions, and orientations.
Secondly, robot motions are not guaranteed to be 100\% reliable and can fail occasionally.
A robot's ability to perform human-like manipulations greatly depends on how it is made; features such as the type of end-effector(s) it has (e.g. what type of gripper it uses, how many fingers it has, etc.), the number of degrees of freedom, its joint types, and the freedom (or lack of) to navigate the environment in search for the items it requires for problem solving.
Through the addition of success rates as weights to FOON, we are better able to capture the uncertainty of performing such actions without failure and to identify a task sequence that is best suited to the current situation.
Furthermore, weights can also be set for robots with different architectures to reflect their ability to perform certain manipulations.

\begin{figure}[t]
	\centering
	\includegraphics[trim={0cm 0.4cm 0cm 0.4cm},clip,width=\columnwidth]{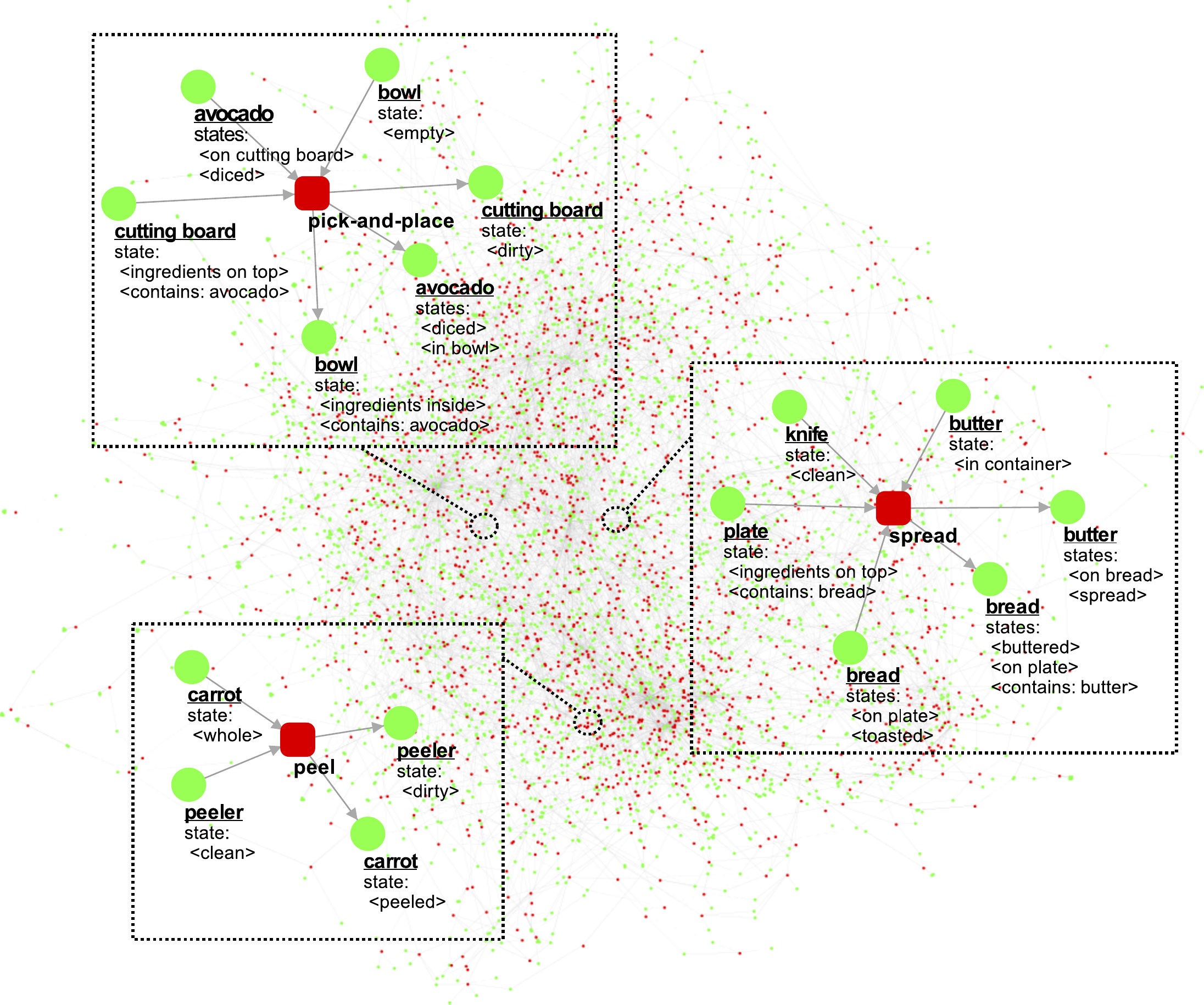}
	\caption{Illustration of our universal FOON made of 100 instructional videos. 
	This network, along with other subgraphs, can be found on our website \cite{foonet}.}
	\label{fig:FOON}
\end{figure}


However, it is important to note that there is no guarantee that a robot will always execute task sequences from FOON with little to no failure.
For instance, an Aldebaran NAO robot (which we use in our experiments) is limited to manipulating lightweight objects, which would require additional effort to design a setting suitable for it to perform tasks; therefore, we address how can a robot use a weighted FOON for problem solving, despite its limited capabilities.
To do so, we simplified the task planning and execution phases as a collaborative robot (cobot) set-up, where we can leverage the abilities and resources available to the robot while introducing collaboration with a human assistant.

This interaction between humans and robots has been extensively explored, where the objective is to achieve a common goal \cite{fong2003survey,yanco2004classifying,goodrich2008human,chandrasekaran2015human} 
by coordinating tasks \cite{mutlu2008robots,randelli2013knowledge,gombolay2015coordination}, in addition to interesting applications such as rehabilitation \cite{robins2004robot,billard2007building} and care for the elderly or disabled \cite{kautz2002overview,rao2002human,dautenhahn2006may}.
In our case with FOON, the human acts as an assistant to the robot, who has the expertise or knowledge to complete the task. 
When given a goal, the robot determines the best course of action through task tree retrieval (where a task tree signifies the sequence of actions that solves the problem) and works with the human to achieve the goal.
This not only makes things easier for both the human person and robot in reducing the complexity of solving the task (in comparison to doing it on their own), but it also improves the overall chance of succeeding in task tree execution.


Our paper is organized as follows: in Section \ref{sec:FOON}, we review the basic structure of FOON, which is followed by Section \ref{sec:retr} where we review how task tree retrieval works.
In Section \ref{sec:HRC}, we introduce a variant retrieval algorithm that considers weights when selecting actions (as functional units) in a task tree while minimizing involvement from a human assistant.
In Section \ref{sec:exp}, we discuss experiments which show that such a system reduces the complexity of task tree execution, even with a robot of limited capabilities.

\section{Functional Object-Oriented Network}
\label{sec:FOON}
\begin{figure}[t]
	\centering
    \includegraphics[width=6.2cm]{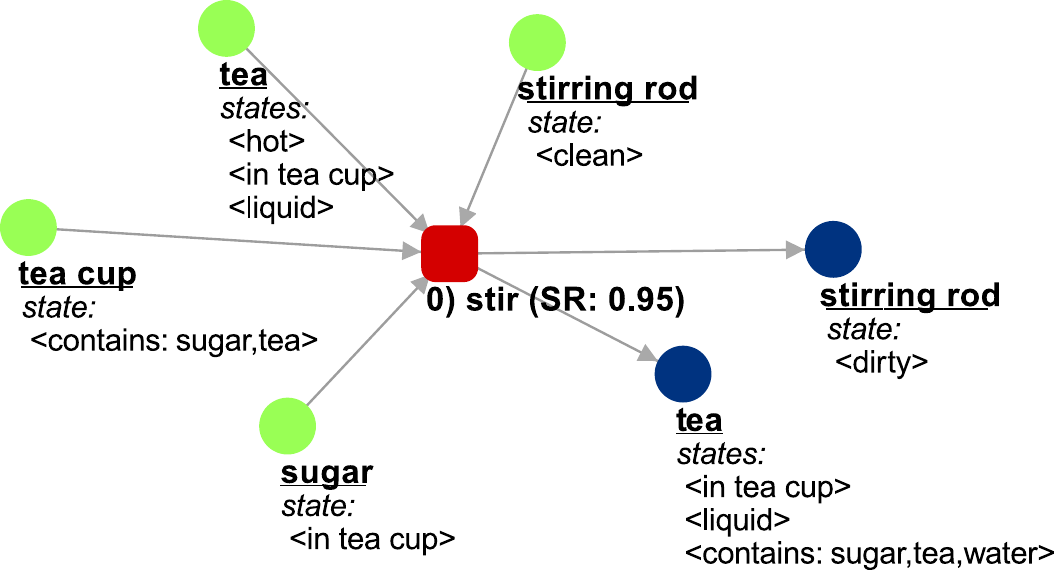}
	\caption{An example of a functional unit (best viewed in colour) with two input nodes (in green) and three output nodes (in dark blue) connected via a motion node (as red square), which describes stirring sugar in tea to sweeten it. 
	The success rate for a certain robot to complete this action is 95\%.
	}
	\label{fig:unit}
\end{figure}

We introduced FOON as a graphical knowledge representation that represents high-level concepts related to human manipulations for service robots.
To represent activities, a FOON has two types of nodes, \textit{object nodes} and \textit{motion nodes}, making FOON a bipartite network.
Affordances are depicted with edges that connect objects to actions, which also indicate order of actions in the network.
The coupling of object and motion nodes to represent a single action is referred to as a \textit{functional unit}.
As an example in Figure \ref{fig:unit}, which describes stirring a cup of tea using a spoon as a functional unit, a {\it spoon} object acts upon a {\it tea cup} object, which contains the ingredients {\it tea} and {\it sugar}; the stirring action is depicted as a motion node with the label {\it stir}.

\subsection{Defining and creating a FOON}
To represent actions in FOON, we denote a collection of object and motion nodes that describe a single action in an activity as a {functional unit}.
A functional unit describes the state change of objects used in a manipulation action before and after execution, where states can be used to determine when an action has been completed~\cite{jelodar2018identifying,jelodar2019joint}.
{\it Input} object nodes describe the required state(s) of objects needed for the task, and {\it output} object nodes describe the outcome of the inputs for said task.
Some actions may not cause a change in all input objects' states, so there may be instances where there are fewer output object nodes than inputs.


A FOON is created using annotations of video demonstrations and converting them into the FOON graph structure; in this annotation process, we note the actions, objects, and state changes (as functional units) that eventually result in a specific meal or product.
A FOON that represents a single activity is referred to as a {\it subgraph}; a subgraph contains functional units in sequence to describe objects' states before and after each action occurs, the time-stamps when each action occurs in the source demonstration, and what objects are being manipulated.
Presently, annotation is done manually by hand, but efforts have been made to investigate how it can be done as a semi-automatic process~\cite{jelodar2018long}.
Two or more subgraphs can be merged together to form a {\it universal} FOON, which is simply a FOON containing information from several sources of knowledge, and as such, a universal FOON could propose variations of recipes.
The merging procedure is simply a union operation done on all functional units from each subgraph we wish to combine; as a result, duplicate functional units are eliminated.
\vspace{-1.5pt}
\subsection{Integrating Weights into a FOON}
In our previous work \cite{paulius2016functional,paulius2018functional}, all motions were treated as having equal weights in a FOON, implying that all manipulations can be executed by any robot without failure and as well as other robots or even humans, as we did not consider the robot's likelihood of successfully executing those actions.
However, this is not realistic since robots are designed and built differently, meaning that they may not all execute manipulations equally in terms of precision and dexterity compared to the original demonstration.

\begin{figure}[t]
	\centering
	\includegraphics[width=\columnwidth]{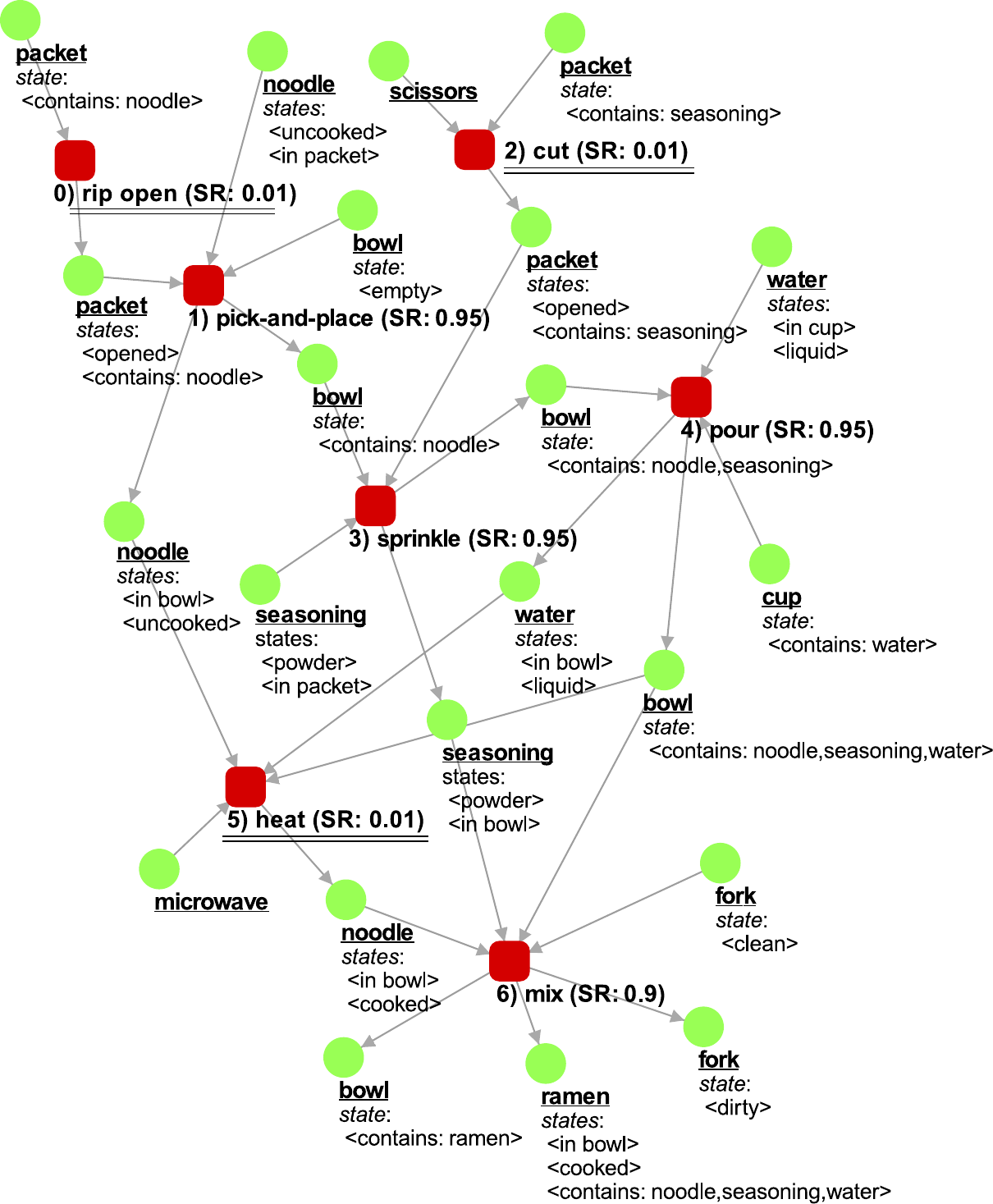}
 	\caption{Illustration of a weighted FOON for making ramen with an overall success rate of 7.716375e-7\%, which is very likely to fail without assistance.}
	\label{fig:tea}
\end{figure}

For these reasons, we introduce weights in FOON to indicate how challenging a manipulation is to perform, where weights reflect the robot's {\it success rate} of performing a given action.
Success rates are assigned as percentages to each functional unit's motion node, and they are not only based on its manipulation type but also the objects in the functional unit.
These values are based on: 1) physical capabilities of the robot, 2) past experiences and ability to execute actions, and 3) the tools or objects to manipulate.
In Figure \ref{fig:tea}, weights ranging between 0 and 1 are assigned to each functional unit.
These weights can be used as heuristics in knowledge retrieval; 
even if several robots are equipped with the same universal FOON (i.e. they share same knowledge from demonstrations), different weights will be assigned to FOON based on the robot's attributes, which can potentially result in very different task trees.
Thus, it is important to note that weights must be defined for each type of robot.
For instance, a small robot like Aldebaran's NAO cannot reliably chop vegetables since it cannot exert the force needed to cut along with lacking the dexterity to do so properly.

We can empirically derive realistic weights for a robot by measuring the frequency of successful trials per manipulation task.
It is important to note that when conducting these experiments, one should vary parameters for these actions (e.g. attributes or sizes of objects) so that one can accurately depict the conditions in which a robot can sufficiently perform those motions.
However, this is not a trivial task, as motions are likely to have many parameters to tune and learn; for example, when learning to scoop with a spoon, several parameters can vary, including where the tool is grasped, the weight of its contents, and the state of matter being scooped. 
Therefore, to simplify this, we assign estimated weights based on our experiences of teaching the robot to perform motions in our experiments.
Motions that cannot be executed by a robot were assigned a success rate of 0.01 (or 1\%), while executable motions were assigned higher values varying between 0.8 and 0.95 (80 to 95\%).
Overall, a robot's ability to perform tasks in FOON is based on its perception, strength, dexterity, and reach within its workspace.

\section{Task Planning with FOON}
\label{sec:retr}
\subsection{Overview of Task Tree Retrieval}
A FOON can not only be used for representing knowledge, but a robot can use it for problem solving.
Given a goal, through the process of task tree retrieval, a robot can obtain a subgraph that contains functional units for actions it needs to execute to achieve it.
A subgraph that is obtained from knowledge retrieval is called a {\it task tree}.
A task tree differs from a regular subgraph, as it will not necessarily reflect a complete procedure from a single human demonstration.
Rather, it will leverage knowledge from multiple sources to produce a novel task sequence.
This search requires a list of items in its environment (i.e. the kitchen) to identify ideal functional units based on the availability of inputs to these units.
This algorithm is motivated by typical graph-based depth-first search (DFS) and breadth-first search (BFS): starting from the goal node, we search for candidate functional units in a depth-wise manner, while for each candidate unit, we search among its input nodes in a breadth-wise manner to determine whether or not they are in the kitchen.
For more details and examples for this algorithm, we refer readers to~\cite{paulius2016functional}.
However, this algorithm does not consider the weights we are now adding to FOON in this work.
Therefore, we will discuss a different approach to find the ideal task tree based on success rates, which would explore all combinations of functional units that can potentially solve the problem.

\begin{algorithm}[t]
\caption{: Path tree retrieval for all possible task trees}
\label{alg:tree}
\begin{algorithmic}[1]
\STATE Let $N_{Goal}$ be the goal object node
\STATE Let $T$ be list of path tree nodes, $R$ be list of roots of $T$ 
\\ \vspace{0.3em}\COMMENT{{\it Find the root functional units for all paths:}}
\FORALL{functional units $FU_{i}$ in $G_{FOON}$}
	\IF {$N_{Goal}$ in $N_{Output}$ of $FU_{i}$}
		\STATE Add $FU_{i}$ to $R$ and $T$ as path tree node
	\ENDIF
\ENDFOR
\\ \vspace{0.3em}\COMMENT{{\it For all path tree roots, build its dependency tree:}}
\WHILE{path tree nodes $t$ exist in $T$}
	\STATE Remove node $t$ from the list $T$
   	\STATE $L_{prelim}$ = \{\}
    \FORALL {$FU_{t}$ in $t$}
    	\FORALL{nodes $N_{Input}$ in $FU_{t}$}
        	\STATE $L_{candidates}$ = \{\}
			\FORALL{functional units $FU_{i}$ in $G_{FOON}$}
				\IF {$N_{Input}$ in $N_{Output}$ of $FU_{i}$}
					\STATE $FU_{candidate}$ = $FU_{i}$
				\ENDIF
                \IF {$FU_{candidate}$ not {\it ancestor}($t$)}
                	\STATE Add $FU_{candidate}$ to $L_{candidates}$
				\ENDIF
			\ENDFOR
            \STATE Add $L_{candidate}$ to $L_{prelim}$
		\ENDFOR
	\ENDFOR
    \\ \vspace{0.3em}\COMMENT{{\it Build path tree nodes for all unit combinations:}}
    \STATE $S_{Cartesian}$ = {\it cartesian\_product}($L_{prelim}$) 
    \FORALL {ordered sets $s$ in $S_{Cartesian}$}
    	\STATE Create new path tree node $t_{new}$ containing $s$
    	\STATE Set parent of $t_{new}$ as current path tree node $t$
        \STATE Add path tree node $t_{new}$ to $T$
    \ENDFOR
\ENDWHILE
\\ \vspace{0.3em}\COMMENT{{\it Perform DFS on $R$ to find all task trees:}}
\FORALL{path tree root nodes $t$ in $R$}
	\STATE Save all paths $P$ found from $DFS(t)$
\ENDFOR
\end{algorithmic}
\end{algorithm}

\subsection{A Weighted Task Tree Retrieval}
The naive algorithm only considers the availability of objects in the robot's environment to determine the best course of action to take to achieve a goal.
However, as with greedy approaches, this algorithm is not likely to find the optimal course of action.
To find the task tree with optimal success rates, it is important to evaluate different paths to a given goal node.
The objective of this algorithm is to build a tree whose nodes can be explored in a depth-wise manner to find the best path of functional units with the best overall success rate.
We refer to these paths as {\it path trees}.

In detail, the algorithm (shown as Algorithm \ref{alg:tree}) works as follows: first, we define a goal node $N_{Goal}$ to the robot.
Path tree root nodes (given in $R$) comprise of an individual functional unit that contains $N_{Goal}$ as output.
Initially, these root nodes are appended to a list of path tree nodes $T$.
For each node $t$ in $T$, we will create and add new tree nodes to $T$ based on their inputs.
We iterate for each input object node (in $N_{Input}$) and identify functional units $FU_{candidate}$ that produce them (i.e. they contain them as output in $N_{Output}$).
For each input, their $FU_{candidate}$ is added to a list $L_{candidates}$, which is then appended to a list $L_{prelim}$ that covers units for all inputs. 
When identifying candidates, we will encounter two cases: there may be several functional units that must be executed in tandem with others to create all necessary input objects (\textit{non-mutually exclusive} events), or there may be multiple units that create each input object to choose from (\textit{mutually exclusive} events).
These can be likened to \textit{AND} and \textit{OR} conditions.
Therefore, path tree nodes of depth 1 or higher can contain more than one functional unit.
Using $L_{prelim}$, we then compute the Cartesian product of candidates to create new path tree nodes for each product set of functional units $S$ in $S_{Cartesian}$, where each set meets input object requirements for $t$.
These new nodes are then added as children of $t$ and appended to $T$.
The connection between a parent and child node stems from the overlapping of a parent's input objects and the child's output objects.
The propagation of path trees continues until we have identified all of the objects needed to solve the problem (i.e., until we can no longer add new leaf nodes).

Once all dependencies are met, we then perform depth-first search to find each individual path $P$ from the root nodes (kept in $R$) to the leaves.
Each path will cover all possible functional units that can be followed to solve the given goal.
The algorithm from \cite{paulius2016functional} will likely produce one of these paths, but as emphasized before, it is not likely to be the optimal path in terms of success rates.
We can use the results from Algorithm \ref{alg:tree} to reduce the search space using available items.
With the inclusion of weights as success rates for each functional unit in FOON, the optimal task tree would be determined by multiplying the robot's success rate for each action (i.e. functional unit) among all path trees.
For example, the total success rate for the sequence in Figure \ref{fig:tea} is equal to 7.716375e-7\%.
Although this is extremely low, we can improve the chance of a robot successfully performing a given task through the assistance of another robot or human.



\begin{figure}[t]
	\centering
	\includegraphics[width=\columnwidth]{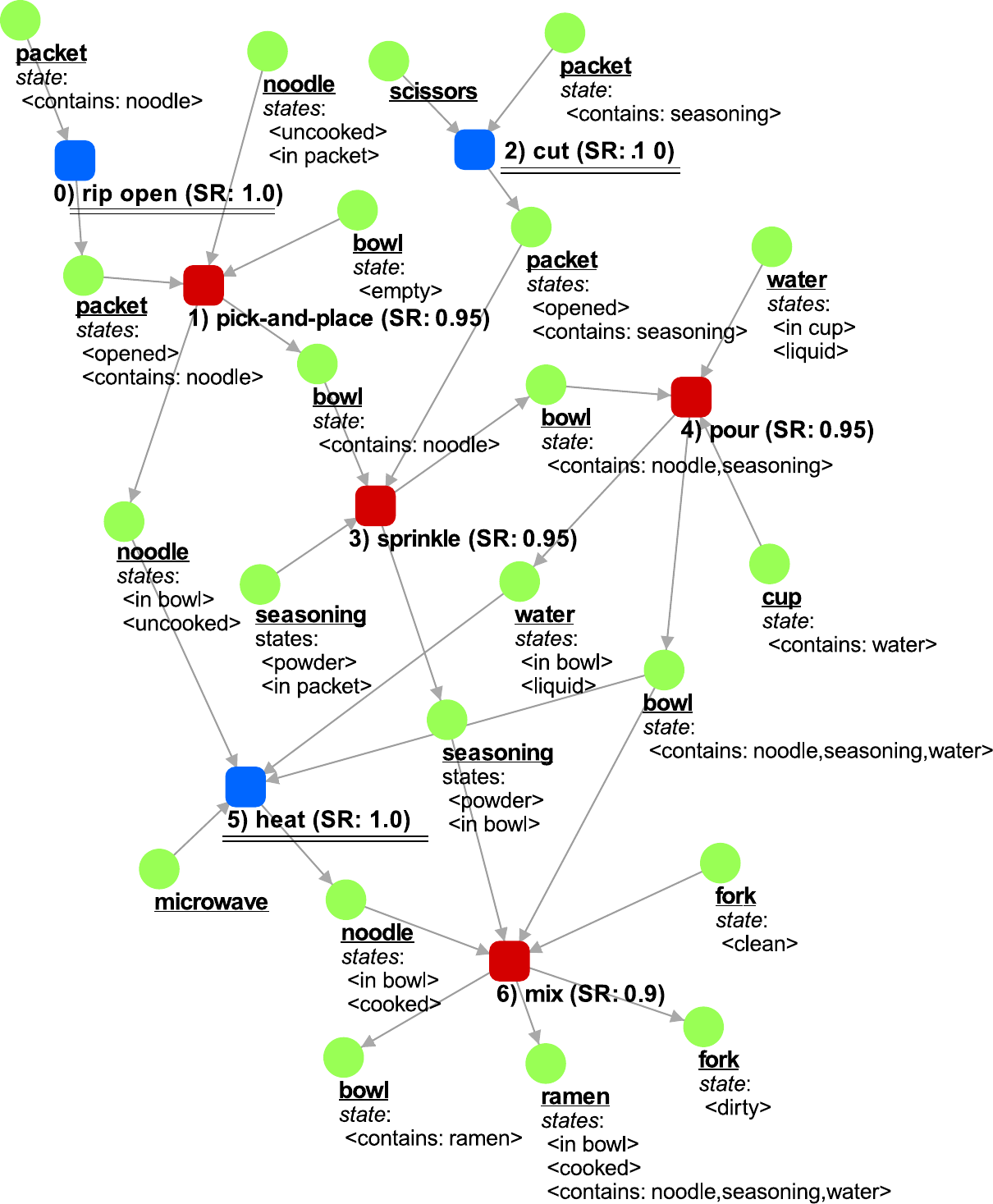}
 	\caption{Illustration of a weighted FOON for making ramen with an overall success rate of 77.16\%, which is attributed to the revised task tree with $M = 3$ such that the robot can succeed in the task with the human's help.}
	\label{fig:tea-2}
\end{figure}

\section{Task Tree Execution with Human Assistance}
\label{sec:HRC}
With the addition of weights reflecting the difficulty of executing motions, we can plan while keeping the robot's capabilities in focus.
However, because certain human motions from demonstrations can be complex, a robot is not guaranteed to perform the same manipulations as well on its own, as it would be difficult to program certain manipulations or perhaps the robot is not built to the task.
Instead of allowing it to act on its own at the risk of failing, it would be best for a robot to work together with assistance to improve its chances of successfully solving the problem.
Assistance can come from another robot or a human assistant who will step in to perform certain actions in its stead.
In this section, we will discuss how manipulations can be executed in a collaborative way with the help of a human assistant.

\begin{figure*}[t]
\centering
	\includegraphics[width=13cm]{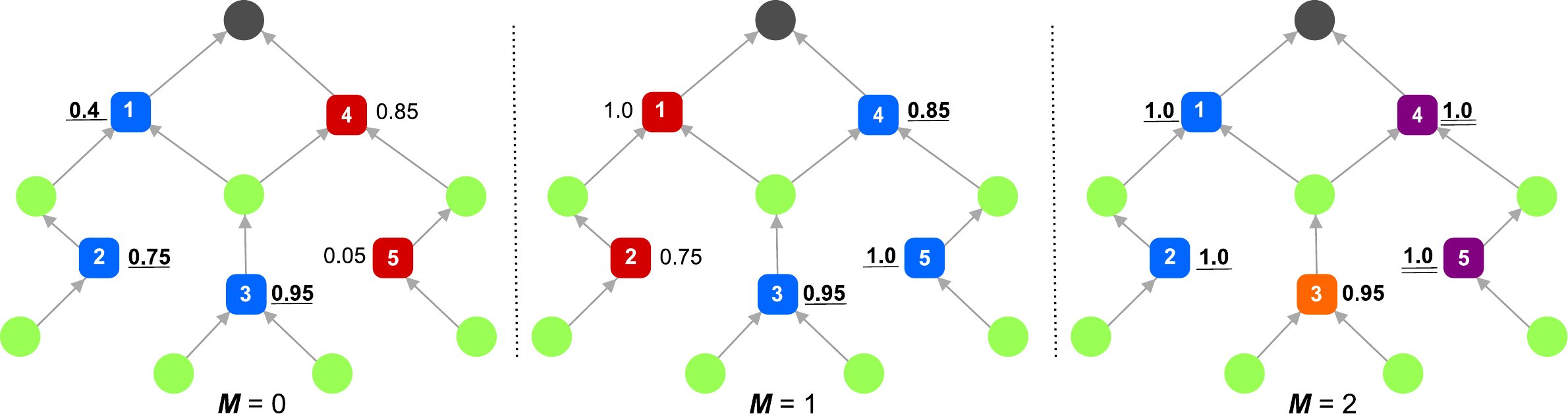}
 	\caption{An example of how task tree retrieval results can change depending on value of $M$ (best viewed in colour). 
    As $M$ changes, the total success rate of each path changes, and thus the ideal task tree will differ. 
    The ideal path tree for the goal (dark grey) is highlighted in blue and underlined. 
    For $M = 0$, the path of functional units \{1, 2, 3\} will be preferred over path \{3, 4, 5\} (28.5\% versus 0.040375\% chance of success); however, for $M = 1$, the path of units \{3, 4, 5\} has a higher weight than the former path (80.75\% versus 71.25\%). 
    When $M = 2$, we can pick either \{1, 2, 3\} or \{3, 4, 5\} as a task tree with a 95\% success rate; here, the two trees are highlighted in blue and purple with a common unit \{3\} highlighted in orange.}
	\label{fig:tree}
\end{figure*}

With the path tree retrieval algorithm, we can obtain novel task trees for different combinations of methods from a universal FOON.
However, certain path trees must be eliminated due to the robot's inability to accomplish the required manipulations for all steps of those task trees, as even the execution of the best task tree can still result in failure.
Aldebaran's NAO robot, for instance, can only manipulate small, light objects; when compared to larger robots such as the PR2 or Baxter, it cannot perform complex manipulations due to its limited workspace and body configuration.
Equally important is its limited mobility to navigate its surroundings since its workspace is very small.
To remedy this, we can involve a human assistant in manipulation problems.
The human assistant, depending on their ability to contribute to the task, can identify the number of steps in a task tree that they are able to perform to cooperatively solve the problem.

As input to the path tree retrieval, the human indicates the number of steps as a value $M$, which cannot exceed the task tree's length $N$ minus 1 step (because if $N$ is equal to $M$, then it means that the human will end up performing the entire task with no robot assistance).
If $M$ is 0, there will be no human involvement in the task. 
The total success rate of a given path $P$ is denoted by the product of all weights among all functional units in $P$, which is simply the joint probability that all actions are successfully executed.
Based on different values of $M$, for each $P$, the success rates will increase by allocating the $M$ lowest units to the human.
For each human-assisted step, the success rate will be set as 100\% for the sake of this paper, unless the human assistant's ability to perform the action is impaired in any way.
It is up to the assistant to determine the degree of involvement they are willing to put into an activity, which realistically depends on factors such as the person's health, mood, and age.
If the human does not provide a value for $M$, the optimal value of $M$ can also be determined by the robot by finding the tree whose success rate at some value of $M$ does not significantly improve over the prior value $M - 1$.
In Figure \ref{fig:tea-2}, the success rate for making ramen noodles increases to 77.16\%, which is high enough to execute to its entirety.
In the task tree execution phase, the robot will perform its delegated actions, and the remaining $M$ steps are given as instructions to the assistant on how to perform them on the robot's behalf.

We illustrate an example as Figure \ref{fig:tree} that shows how candidate task trees are weighed against one another and how the total success rate can change between a pair of trees when there is human involvement.
As the value of $M$ becomes higher, the ideal task tree changed within trees and caused a significant improvement in the overall success rate of the task (from 28.5\% to 95\%).
However, we can probably make a reasonable trade-off with $M = 1$ instead of $M = 2$, since it should demand less effort from the human assistant.

\begin{figure}[t]
	\centering
	\includegraphics[width=7cm]{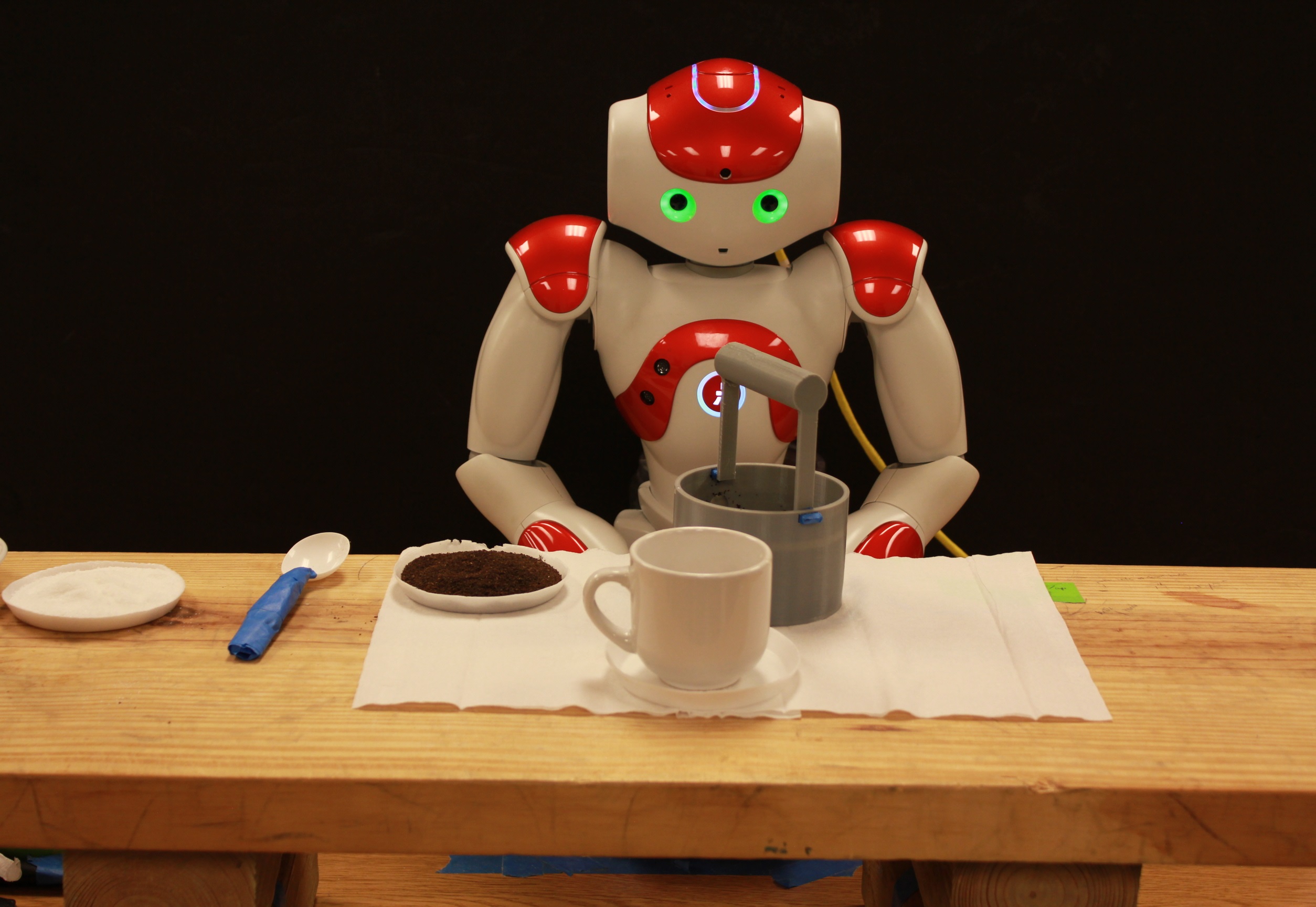}
 	\caption{Experimental setup with the NAO robot for our weighted FOON demonstration. Its primitives are taught via kinesthetic teaching.}
	\label{fig:nao}
\end{figure}

\begin{figure*}[ht]
\centering
	\includegraphics[width=17cm]{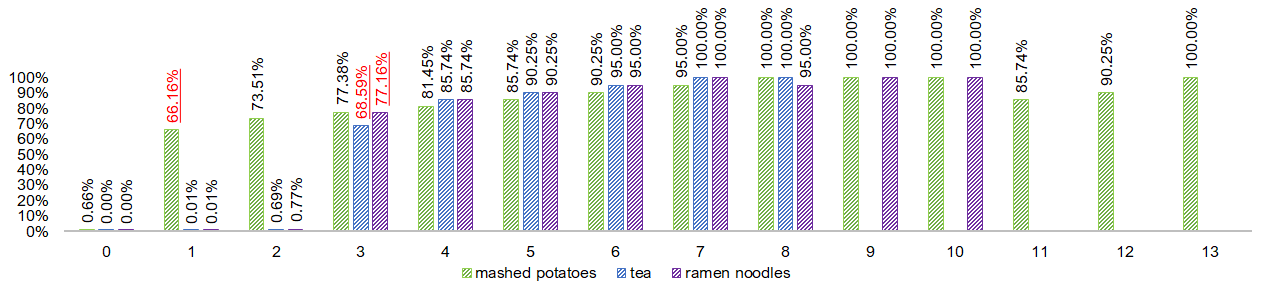}
 	\caption{Graph showing the gradual improvement in success rates (y-axis) as $M$ (x-axis) increases. 
    Sudden drops between $M$ signifies that other paths are considered that exceed the length of $M$, resulting in a completely human tree (e.g. for values $M$ = 10 and $M$ = 11, the best potential path trees are different in length).
    Bars are omitted for values of $M$ that exceed the length of a task tree.
    The values in red indicate the path tree used in Section \ref{sec:exec}. }
	\label{fig:SR}
\end{figure*}

\section{Experiments}
\label{sec:exp}
In our experiments, our aim was to show that we can improve a robot's performance through collaboration in the task execution phase.
To demonstrate this, we want to show that a robot can acquire the ideal task tree for execution, delegate commands to the human assistant, and successfully obtain the goal product at varying degrees of involvement.
We used the NAO robot to execute manipulations for the tasks of making tea, mashed potatoes, and ramen noodles.
Variations of preparing each dish were merged together into a universal FOON, which was then provided to the algorithm to identify different paths to prepare these items and to show how functional units are selected based on success rates.
Because the NAO robot is very small, its physical capabilities are limited to using smaller versions of items, and furthermore, certain manipulations were very difficult to replicate.
Under these circumstances, the robot can greatly benefit from human participation in the task tree execution phase.  
Certain parts of the tasks, such as heating to obtain hot water, cannot be performed by the robot; those motion nodes were assigned a very low success rate of 1\% to reflect how impossible they are for the robot to do on its own.
However, for robot-executable motions, we assigned higher rates based on our confidence in the robot performing the programmed motion primitives.
The task trees obtained via retrieval and demonstrations of the robot executing the final trees with a human assistant are provided as supplementary material\footnote{Link to Material -- \url{http://foonets.com/human-robot.html}}.

\subsection{Finding the optimal task tree for NAO}
\label{sec:find-tree}
Firstly, we show that we can obtain optimal task trees suitable for the NAO robot to prepare tea, mashed potatoes, and ramen noodles.
To improve the overall success rate of each activity, the algorithm is expected to iterate through several values of $M$ to determine the ideal $M$ that balances the effort required from the robot and the human assistant.
We show the best overall success rates in the graph as Figure \ref{fig:SR} to show how success rates increased as $M$ increases; the chances of success significantly improve as more steps are delegated to the human assistant. 
Based on the success rates assigned to the NAO robot's universal FOON, the values of $M$ that were ideal for balanced human-robot manipulations were $M = 1$, $M = 3$, and $M = 3$ for the tasks of mashed potatoes, tea-making, and ramen noodles respectively (whose success rates are highlighted in red in Figure \ref{fig:SR}).
Although some primitives have questionably low success rates, it will still be able to execute the task tree.
Within the supplementary material, the task trees have the same number of units labelled as ``human-executable" as $M$.  

\subsection{Executing the optimal task trees}
\label{sec:exec}
Secondly, we show that we can successfully perform these actions as a cobot task.
The NAO robot was programmed to modularly execute certain motions as described by a task tree's object and motion nodes. 
Since the objective of this work was to demonstrate the use of a universal FOON in task planning, each motion skill/primitive that could be executed by the robot (e.g. pouring, stirring) were taught by manually recording trajectories via kinesthetic teaching to simplify programming the robot and to reduce the complexity of the problem space.
We did not apply any sensors nor vision systems for manipulation, as there was no need for object detection at this stage.
The order of executing each functional unit was determined by the order in which they were sequenced in the acquired task tree; in other words, the NAO robot simply follows each step and executes them one by one when able.
In the supplementary material, we provide video demonstrations of the execution of manipulations shown in each task tree and show how they were carried out based on $M$.
Without human assistance ($M = 0$), the NAO robot attempted to execute the task tree which ended in failure once it encountered the motion it did not know how to perform (with success rate of 1\%); however, with human assistance, the robot finished the tasks and made the final product.
In some cases, we did observe that the motion primitives of the robot can fail, rendering the entire sequence as a failure.
To overcome this through human interaction, the robot can request feedback from the assistant to determine if it should perform the action again.
As future work, to better solve this, we will include sensors or recovery protocols so that it knows when it has failed a particular action and explore how to recover from failure.




\section{Conclusion and Future Work}
To summarize, we proposed an improved functional object-oriented network (FOON) with weights for cobot task planning and execution.
Such weights reflect a robot's ability to perform actions, which are described as functional units in a FOON.
With a retrieval procedure that takes a robot's physical capabilities into account, we can determine if it can successfully execute the task tree on its own or if it requires assistance.
If a robot cannot execute an entire sequence of manipulations on its own, a human assistant can perform difficult motions with guidance from a robot.
In our experiments, we show that we can obtain suitable task trees that leverage both the robot's and human's capabilities while minimizing effort from the human.

In the future, we will further explore cobot planning and focus on important aspects that we did not consider in this paper, such as recovery from failure or demonstrating skills when an unknown manipulation type is encountered.
Further, rather than human assistance, we will explore task tree execution between multiple robots, even of different types, for workload sharing. 
In this way, resources and tasks will be distributed among robots to achieve a common goal.


\section*{Acknowledgement}
\noindent This material is based upon work supported by the National Science Foundation under Grant Nos. 1910040 and 1812933.

\bibliographystyle{unsrt}
\bibliography{ref}

\end{document}